\documentclass[letterpaper,10pt,conference]{ieeeconf}  
\IEEEoverridecommandlockouts                              
\usepackage{amsmath}
\usepackage{amssymb}
\usepackage{graphicx}
\usepackage{cite}
\usepackage{comment}
\usepackage{booktabs}
\usepackage{float}
\usepackage{xcolor}
\usepackage{multirow}
\usepackage{array}
\usepackage{algorithm}
\usepackage{algpseudocode}

\title{\LARGE \bf Collaborative System Failure Prognostics via Federated Longitudinal–Survival Modeling}

\author{Fan Yang$^{1}$,
Madelyn Weller$^{1}$,  Dimuthu Fernando$^{2}$, Hila Livneh$^{1}$, and Yuxin Wen$^{1}$ 
\thanks{*This work was supported in part by the National Science Foundation under Grant Nos. 2501643 and 2518849, and in part by The Fletcher Jones Foundation.}
\thanks{$^{1}$ F. Yang,  M. Weller,  H. Livneh and Y. Wen are with the Fowler School of Engineering, Chapman University, Orange, CA 92866 USA
        {\tt\small fayang@chapman.edu, mweller@chapman.edu, livneh@chapman.edu, yuwen@chapman.edu}}%
\thanks{$^{2}$D. Fernando  is with the Department of Statistics, Grand Valley State University, Allendale, MI 49401 USA
        {\tt\small fernadim@gvsu.edu}}%
}

\begin{document}

\maketitle
\thispagestyle{empty}
\pagestyle{empty}

\begin{abstract}
Time-to-event modeling provides a systematic framework for estimating time-dependent failure risk, reliability, and remaining useful life (RUL) from longitudinal condition monitoring data. However, applying these models to distributed prognostics remains challenging as  sensor trajectories and failure-time records are often stored across organizations or operational sites and cannot be centrally pooled due to privacy or proprietary constraints. Moreover, the classical Cox proportional hazards model relies on a nonseparable partial likelihood involving global risk sets, making direct optimization difficult under standard federated learning protocols. This paper presents a federated longitudinal–survival modeling framework for collaborative system failure prognostics. The proposed framework combines longitudinal sensor representation learning with a client-separable discrete-time hazard objective, enabling multiple clients to collaboratively train a prognostic model without sharing raw sensor measurements or individual failure records. Time-dependent representations extracted from multivariate sensor histories are used to estimate interval-specific failure hazards, reliability curves, and RUL of systems. Experiments on the four C-MAPSS turbofan engine degradation subsets under simulated decentralized settings demonstrate that the proposed framework consistently improves prognostic performance over isolated local training while maintaining performance comparable to centralized training across heterogeneous operating conditions and failure modes. These results demonstrate the potential of federated longitudinal–survival modeling for collaborative, data-aware condition monitoring and system failure prognostics.
\end{abstract}

\section{Introduction}

Maintaining high reliability in complex mechanical systems is essential for ensuring operational efficiency and mission success in demanding environments \cite{lei2018machinery}. Failure prognostics plays a central role in this objective by estimating future failure risk and remaining useful life (RUL), thereby enabling proactive maintenance before systems violate safety-critical operating specifications. Recent advances in the Industrial Internet of Things (IoT) and onboard sensing technologies have made it possible to continuously monitor system health through high-fidelity, multivariate sensor streams. These data provide detailed longitudinal records of degradation processes and create new opportunities for dynamic failure prediction in complex engineering systems \cite{zio2022prognostics,wen2022recent}. Given such degradation histories, system failure prognostics seeks to answer fundamental maintenance questions, including: `What is the probability that a system will continue operating beyond a future time \(t\)?' and `How much useful life remains before failure?'. Reliable answers to these questions are critical for predictive maintenance, operational risk assessment, and maintenance scheduling.

Over the past decade, numerous failure prediction models have been developed to support informed prognostics and maintenance decision-making in complex systems \cite{zhu2019survey, wen2018degradation, zio2022prognostics, wen2022recent}. These models generally fall into two categories: soft failure models, which define failure as degradation exceeding a predefined operational threshold, and hard failure models, which characterize sudden, catastrophic events using hazard functions. Soft failure models leverage longitudinal sensor data, such as vibration, temperature, and pressure signals, to capture the temporal evolution of system health. In contrast, hard failure models rely on time-to-event data, typically derived from maintenance logs, to estimate the probability of discrete failure events. In practice, these failure modes are often intrinsically correlated. For example, in hydraulic actuators, wear-induced degradation (soft failure) can increase debris accumulation, subsequently increasing the likelihood of abrupt clamp stagnation (hard failure) \cite{hu2020condition}. By integrating longitudinal sensor streams with time-to-event data, a more comprehensive representation of system health can be achieved. Existing models that focus exclusively on one failure type may ignore these underlying correlations, potentially leading to suboptimal reliability estimates in safety-critical deployments.

To leverage both degradation trajectories and failure-time information, recent studies have explored joint modeling frameworks for the simultaneous analysis of longitudinal and time-to-event data. For example, Zhou et al. applied joint modeling to failure prediction in automotive lead-acid batteries \cite{zhou2014remaining}. Later, Yan et al. used functional principal component analysis (FPCA) within a joint modeling framework to learn longitudinal degradation trajectories and improve dynamic prediction \cite{yan2018functional}. Yue et al. combined multivariate Gaussian convolution processes with the Cox proportional hazards (PH) model to fuse time-series observations and survival outcomes for event prediction \cite{yue2021joint}. Wen et al. further introduced a neural-network-based Cox model for IoT signal fusion and failure prediction, showing how deep representations can be integrated with hazard-based prognostics \cite{wen2023neural}. More recently, Brumm et al. developed a joint modeling framework for degradation signals and time-to-event data that uses particle filtering for online learning and dynamic RUL prediction \cite{brumm2025joint}. A key advantage of these survival-based formulations is that they do not require a pre-specified degradation threshold.  Instead, failure risk is modeled through the hazard function and the corresponding survival probability.

The joint models above rely largely on parametric or kernel-based degradation representations. Deep learning has  expanded modeling capacity along two largely separate lines. Neural survival models, such as DeepSurv and DeepHit \cite{katzman2018deepsurv,lee2018deephit}, learn nonlinear risk functions or event-time distributions. However, they operate on fixed baseline covariates and  do not accommodate the time-varying degradation state that prognostics requires. In Prognostics and Health Management (PHM), convolutional, recurrent, attention-based, and Transformer-based models learn degradation representations from multivariate sensor streams for RUL prediction \cite{sateesh2016deep,li2018remaining,zhang2022dast}, but most formulate the task as direct RUL regression rather than survival learning, returning a point estimate of remaining life without hazard functions, survival probabilities, or time-dependent failure risks.

These modeling advances assume access to a single, sufficiently rich training corpus. In practice, system-level prognostic data, including sensor histories and corresponding failure-time outcomes, are often limited within a single manufacturer, fleet operator, or maintenance center. As a result, locally trained models may not capture sufficient variation in degradation patterns and failure mechanisms. Collaborative learning across organizations can improve prognostic reliability by leveraging broader operational experience, but direct data pooling is often infeasible due to privacy, proprietary, and operational constraints \cite{zhang2023privacy}. Federated learning (FL) provides a natural paradigm for decentralized model training, with FedAvg enabling collaborative optimization without centralizing raw data \cite{mcmahan2017communication,kairouz2021advances}. Extensions such as FedProx and SCAFFOLD further address client heterogeneity and client drift in distributed optimization \cite{li2020fedprox,karimireddy2020scaffold}. Recent federated prognostics studies have also explored collaborative RUL estimation across distributed industrial assets, including FedAvg-based failure prognosis and feature-matched aggregation for heterogeneous edge devices \cite{dhada_empirical_2020,arunan2023federated}.

Extending federated learning from RUL regression to time-to-event modeling, however, raises a challenge that standard federated optimization does not resolve. Many joint longitudinal--time-to-event models use the Cox proportional hazards (PH) model to relate an evolving degradation state to failure risk. Although the Cox model accommodates time-dependent system conditions without requiring a fully specified failure-time distribution \cite{cox1975partial}, its partial likelihood depends on globally ordered failure times and risk sets that may span multiple clients. The objective is therefore not client-separable, and averaging locally computed partial-likelihood gradients does not recover the global objective.  Only a limited number of studies have addressed this through distributed Cox optimization or discrete-time reformulations \cite{andreux2020federated,zhang2022federated}. These methods, however, were developed for static, low-dimensional clinical covariates and evaluate risk ranking, leaving federated time-to-event modeling for high-dimensional longitudinal sensor data largely unexplored.

Building on this emerging research, we develop a federated time-to-event prognostic framework for complex engineering systems with high-dimensional sensor streams. To make survival learning compatible with federated optimization, we use a discrete-time formulation whose interval-level likelihood decomposes into a sum of client-local terms, removing the global risk sets that couple the continuous-time partial likelihood across clients. Within this formulation, we learn representations of longitudinal sensor trajectories through a temporal encoder and adopt the complementary log-log link to obtain an exact grouped-data proportional-hazards model. Interval-specific failure hazards, reliability curves, and RUL are then estimated from local sensor histories without sharing raw sensor trajectories or failure-time records.
The main contributions of this work are as follows:
(1) We integrate high-dimensional, time-dependent sensor trajectories with failure-time outcomes in a unified prognostic framework, connecting longitudinal degradation representation learning with hazard-based failure-risk estimation.
(2) We adapt a discrete-time Cox formulation to the federated setting, showing that its interval-level likelihood is client-separable and therefore eliminates global risk-set construction during local optimization and parameter aggregation.
(3) We evaluate the proposed framework on the four C-MAPSS high-fidelity turbofan engine degradation datasets under decentralized settings. The results demonstrate that collaborative training consistently improves prognostic performance over isolated local training while retaining strong performance relative to centralized data pooling.

The remainder of the paper is organized as follows. Section \uppercase\expandafter{\romannumeral2} presents the proposed methodology. In Section \uppercase\expandafter{\romannumeral3}, a case study is presented to investigate the effectiveness of the proposed method. Section \uppercase\expandafter{\romannumeral4} concludes the paper and outlines future research directions.

\section{Methodology}
\label{sec:methodology}
This section presents the proposed privacy-aware prognostic framework. First, the federated learning setting is defined to formalize collaborative model training under data-local constraints. Next, the continuous-time Cox PH model is introduced, and the limitation imposed by its nonseparable partial likelihood is addressed through a discrete-time reformulation suitable for federated optimization. Finally, longitudinal sensor histories are transformed into degradation representations and integrated with the resulting client-separable hazard model to estimate failure risk and RUL.

\subsection{Federated Learning Setting}
Federated learning enables multiple clients to train a shared model without exchanging raw data. This setting is relevant to industrial prognostics, where sensor trajectories and failure records may be distributed across maintenance centers, manufacturers, or fleet operators and subject to proprietary or operational constraints \cite{dhada_empirical_2020}.

Considering a set of $K$ clients, each client represents an independent  site. 
Each site $k$ operates multiple systems and maintains a local dataset $\mathcal{D}_k = \{ (\mathbf{X}_i, T_i, \delta_i) \}_{i \in \mathcal{N}_k}$, where $\mathcal{N}_k$ denotes the index set of systems belonging to client $k$.
For each system $i$, $\mathbf{X}_i = [X_{i1}, X_{i2}, \dots, X_{iT_i}]$ represents the multivariate longitudinal sensor trajectory observed up to time $T_i$, with $T_i$ denoting the failure or censoring time and $\delta_i \in \{0,1\}$ is the corresponding event indicator, with $\delta_i = 1$ if system $i$ experiences the failure event and $\delta_i = 0$ if the observation is right-censored.
The global objective is to optimize a shared model parameter $\boldsymbol{\Theta}$ by aggregating local learning processes. The global optimization objective in the federated setting is formulated as the weighted aggregation of client-specific losses:
\begin{equation}
\mathcal{L}_{F}(\boldsymbol{\Theta}) 
= \sum_{k=1}^{K} \frac{N_k}{N} \, \mathcal{L}_{k}(\boldsymbol{\Theta}),
\end{equation}
where $N_k = |\mathcal{N}_k|$ denotes the number of systems maintained by client $k$, and 
$N = \sum_{k=1}^{K} N_k$ is the total number of systems across all clients. Federated learning aims to minimize $\mathcal{L}_{F}(\boldsymbol{\Theta})$ through iterative collaboration without exchanging raw data. 
 At each communication round $r$, the server broadcasts $\boldsymbol{\Theta}^{(r)}$ to the clients. Each client performs local optimization on $\mathcal{D}_k$ and returns the updated parameters $\boldsymbol{\Theta}_{k}^{(r+1)}$. The server then applies weighted averaging,
\begin{equation}
\boldsymbol{\Theta}^{(r+1)}
=
\sum_{k=1}^{K}
\frac{N_k}{N}
\boldsymbol{\Theta}_{k}^{(r+1)}.
\label{eq:fedavg}
\end{equation}
This procedure is repeated until convergence or until a predefined stopping criterion is satisfied.

\subsection{Continuous-Time Cox PH Model}
Let $T$ denote a nonnegative event time and $\mathbf{z}\in\mathbb{R}^{d}$ a covariate vector. The continuous-time hazard function is
\begin{equation}
\lambda(t\mid\mathbf{z})
=
\lim_{\Delta t\rightarrow 0}
\frac{
\Pr\left(t\leq T<t+\Delta t\mid T\geq t,\mathbf{z}\right)
}{\Delta t}.
\label{eq:continuous-hazard-def}
\end{equation}
The Cox PH model specifies
\begin{equation}
\lambda(t\mid\mathbf{z})
=
\lambda_{0}(t)\exp\left(\boldsymbol{\beta}^{\top}\mathbf{z}\right),
\label{hazard_eq}
\end{equation}
where $\lambda_{0}(t)$ is an unspecified baseline hazard and $\boldsymbol{\beta}$ is a coefficient vector. For two covariate vectors $\mathbf{z}_1$ and $\mathbf{z}_2$, the hazard ratio is
\begin{equation}
\frac{\lambda(t\mid\mathbf{z}_1)}
{\lambda(t\mid\mathbf{z}_2)}
=
\exp\left[\boldsymbol{\beta}^{\top}(\mathbf{z}_1-\mathbf{z}_2)\right],
\end{equation}
which is constant over time under the PH assumption.

The regression coefficients are commonly estimated by maximizing the partial log-likelihood
\begin{equation}
\ell(\boldsymbol{\beta})
=
\sum_{i:\delta_i=1}
\left[
\boldsymbol{\beta}^{\top}\mathbf{z}_i
-
\log\left
\{
\sum_{q\in\mathcal{R}_i}
\exp\left(\boldsymbol{\beta}^{\top}\mathbf{z}_q\right)
\right\}
\right],
\label{eq3}
\end{equation}
where $\mathcal{R}_i=\{q:T_q\geq T_i\}$ is the risk set at event time $T_i$. The denominator couples each event contribution to all observations in the corresponding risk set. When the observations are distributed across clients, direct evaluation of Eq.~\eqref{eq3} requires global risk-set information and is therefore not naturally compatible with standard federated averaging. The fixed-covariate formulation in Eq.~\eqref{hazard_eq} also does not directly encode the high-dimensional longitudinal patterns used for dynamic prognostics. Although the counting-process extension of the Cox model accommodates time-varying covariates \cite{andersen1982cox}, its partial likelihood remains defined through event-specific risk sets that couple contributions across observations and, in distributed settings, across clients \cite{cox1975partial,andreux2020federated}. A federated prognostic formulation must therefore support learned temporal representations while avoiding global risk-set construction.

\subsection{Federated Longitudinal--Survival Framework}
\begin{figure*}[t]
    \centering
    \vspace{-0.7cm}
    \includegraphics[width=\textwidth]{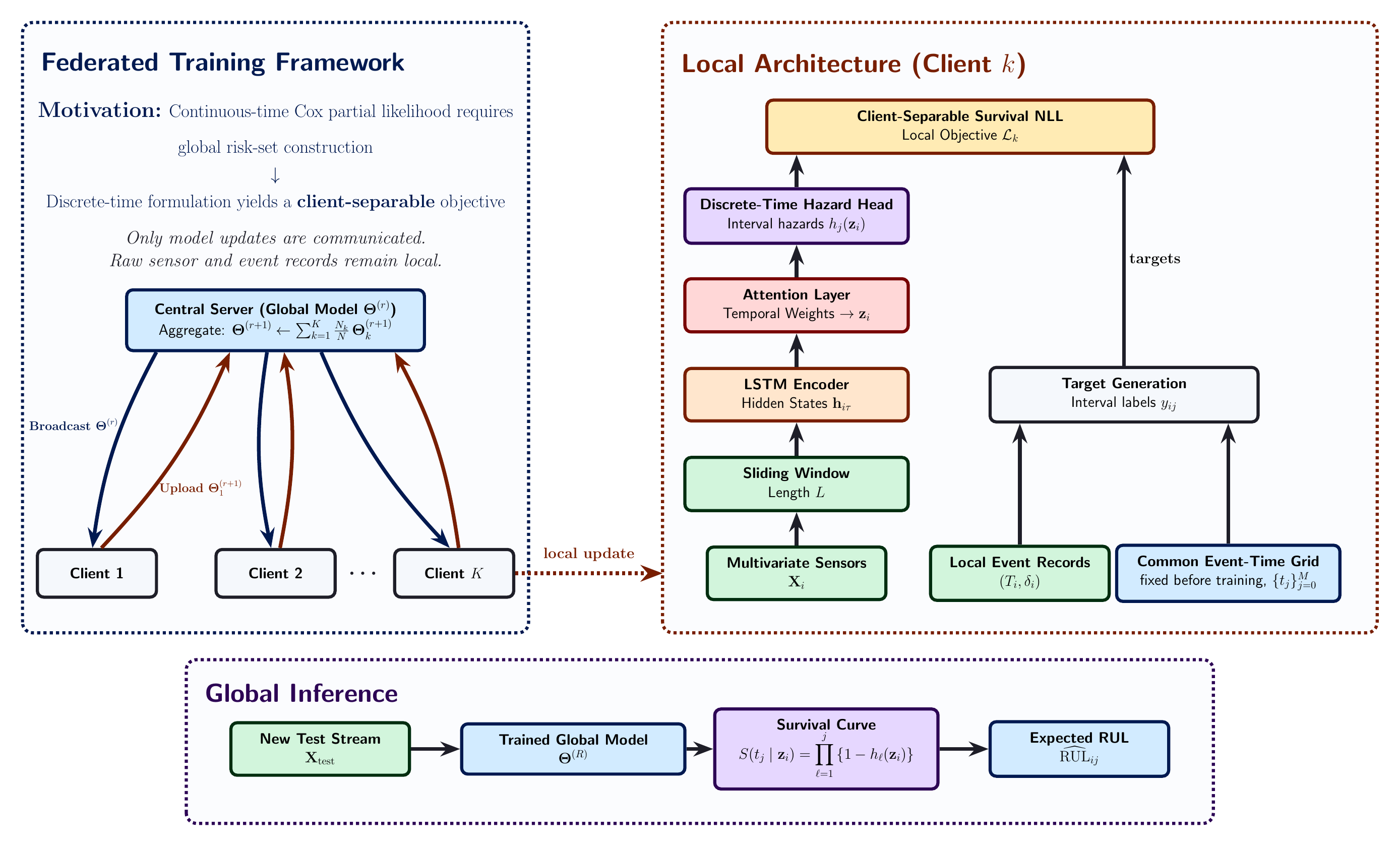}
    \vspace{-0.77cm}
    \caption{Overview of the federated longitudinal--survival framework. 
    A common event-time grid is used to construct interval labels over the locally observed at-risk intervals. Each client trains an LSTM-attention encoder and a discrete-time hazard model using its local sensor windows and failure records, and only model updates are transmitted for federated aggregation. The trained global model produces interval-specific failure risks, reliability curves, and RUL estimates for new sensor windows.}
    \vspace{-0.3cm}
    \label{fig:framework_overview}
\end{figure*}

A federated longitudinal--survival framework is developed for collaborative prognostics using decentralized sensor data. Building on the federated discrete-time survival formulation of \cite{andreux2020federated}, in which the interval-level objective is separable across subjects, the proposed framework learns failure hazards collaboratively without global risk-set construction. Within each client, a temporal encoder maps longitudinal sensor trajectories to representations that drive the estimation of interval-specific failure hazards. These hazard estimates support dynamic prediction of failure risk without sharing raw sensor data or individual failure records. Fig.~\ref{fig:framework_overview} provides an  overview of the proposed framework. The individual components are described in the following subsections.

\subsubsection{Time Discretization and Client-Separable Hazard Learning}
\label{sec:Discretization}
The continuous-time Cox partial likelihood is non-separable because each event contribution depends on a risk set $\mathcal{R}_i$ that may span multiple clients. Consequently, Eq.~\eqref{eq3} cannot be decomposed into independent client-local objectives without communicating global risk-set information.
To obtain a client-separable objective, we reformulate the survival problem over a finite set of non-overlapping time intervals. Let $0=t_0<t_1<\cdots<t_M$ denote a discretization grid, where the horizon $t_M$ is
specified from domain knowledge, such as the monitoring window or the maximum operating
life of the system, so that $T_i\in(0,t_M]$ for all systems. The $j$th interval is
$
I_j=(t_{j-1},t_j],  j=1,\ldots,M.
$
The left-open, right-closed convention assigns each observed time to a unique interval so that an event occurring exactly at $t_j$ falls in $I_j$ rather than $I_{j+1}$. Let
$J_i\in\{1,\ldots,M\}$ denote the index of the interval containing $T_i$.
The observed time of each system is thereby converted into a sequence of $J_i$
interval-level binary labels, one for each interval in which the system remains under
observation:
\begin{equation}
y_{ij} =
\begin{cases}
1, & \text{if } j=J_i \text{ and } \delta_i=1,\\
0, & \text{otherwise},
\end{cases}
\qquad j=1,\ldots,J_i,
\label{eq:stacked-label}
\end{equation}
where $\delta_i=1$ indicates failure and $\delta_i=0$ censoring. Thus, a failed system has a sequence of zeros followed by a one in its failure interval, whereas a censored system contributes only zeros up to its censoring
interval.

For each interval $I_j$, the discrete-time hazard is defined as the
conditional probability that system $i$ fails within this interval, given that it
has survived to the beginning of the interval and given its covariate or feature
information $z_i$:
\begin{equation}
h_j(\mathbf{z}_i)
=
\Pr\left(T_i\in I_j\mid T_i>t_{j-1},\mathbf{z}_i\right),
\label{eq:discrete-hazard-def}
\end{equation}
where $\mathbf{z}_i$ is the representation extracted from the sensor trajectories available at the prediction time. Unlike the continuous-time hazard $\lambda(t\mid\mathbf{z}_i)$, $h_j(\mathbf{z}_i)$ is a conditional failure probability over a finite interval.

To connect Eq.~\eqref{eq:discrete-hazard-def} with the continuous-time Cox model, let
$S(t\mid\mathbf{z}_i)$ denote the conditional survival function, representing the probability that system $i$ remains operational beyond time $t$. Under Eq.~\eqref{hazard_eq},
\begin{equation}
S(t\mid\mathbf{z}_i)
=
\exp\left[-\int_{0}^{t}\lambda(u\mid\mathbf{z}_i)\,du\right]
=
S_0(t)^{\exp(\boldsymbol{\beta}^{\top}\mathbf{z}_i)},
\label{eq:cont-survival}
\end{equation}
where
$S_0(t)=\exp[-\int_{0}^{t}\lambda_0(u)\,du]$.
The conditional probability of surviving interval $I_j$ is
\begin{equation}
1-h_j(\mathbf{z}_i)
=
\Pr\left(T_i>t_j\mid T_i>t_{j-1},\mathbf{z}_i\right)
=
\frac{S(t_j\mid\mathbf{z}_i)}
{S(t_{j-1}\mid\mathbf{z}_i)}.
\label{eq:surv-ratio}
\end{equation}
Let
\begin{equation}
\Lambda_{0j}
=
\int_{t_{j-1}}^{t_j}\lambda_0(u)\,du
\label{eq:baseline-cumhaz}
\end{equation}
denote the baseline cumulative-hazard increment over $I_j$. Substitution into Eq.~\eqref{eq:surv-ratio} gives
\begin{equation}
1-h_j(\mathbf{z}_i)
=
\exp\left[
-\Lambda_{0j}
\exp\left(\boldsymbol{\beta}^{\top}\mathbf{z}_i\right)
\right].
\label{eq:discrete-surv-ratio}
\end{equation}
Applying the complementary log-log transformation yields
\begin{equation}
\operatorname{cloglog}\{h_j(\mathbf{z}_i)\}
=
\log\{-\log[1-h_j(\mathbf{z}_i)]\}
=
\alpha_j+\boldsymbol{\beta}^{\top}\mathbf{z}_i,
\label{eq:discrete-cloglog2}
\end{equation}
where
$
\alpha_j=\log\Lambda_{0j}.
\label{eq:baseline-alpha}
$
Equivalently,
\begin{equation}
h_j(\mathbf{z}_i)
=
1-
\exp\left\{
-\exp\left(
\alpha_j+\boldsymbol{\beta}^{\top}\mathbf{z}_i
\right)
\right\}.
\label{eq:discrete-linear-hazard}
\end{equation}
For a reference instance satisfying
$\boldsymbol{\beta}^{\top}\mathbf{z}_i=0$,
Eq.~\eqref{eq:discrete-linear-hazard} reduces to
$
h_j(\mathbf{z}_i)
=
1-\exp\left\{-\exp(\alpha_j)\right\}.
$
Hence, $\alpha_j$ represents the log baseline cumulative-hazard increment over interval $I_j$. If the covariates are treated as constant within each interval, discretizing the continuous-time Cox model yields the complementary log-log hazard formulation directly \cite{prentice1978regression,kalbfleisch2002statistical}. Therefore, the cloglog link function preserves the proportional-hazards interpretation of the original Cox model after discretization. An alternative commonly used in discrete-time survival analysis is the logit link function, which corresponds to a proportional-odds formulation \cite{allison1982discrete}. The cloglog and logit links yield similar hazard estimates when interval-specific failure probabilities are small. We adopt the cloglog link as it preserves the connection to the continuous-time Cox PH model.

Under the formulation, the survival process is represented as a sequence of
conditional Bernoulli trials over the at-risk intervals. Thus, the likelihood
contribution of system $i$ is
\begin{equation}
L_i=\prod_{j=1}^{J_i}
h_j(\mathbf{z}_{i})^{y_{ij}}
\left(1-h_j(\mathbf{z}_{i})\right)^{1-y_{ij}}.
\label{eq:instance-likelihood}
\end{equation}
The corresponding mean negative log-likelihood at client $k$ is
\begin{equation}
\begin{aligned}
\mathcal{L}_k(\boldsymbol{\Theta})
=
-\frac{1}{N_k}
\sum_{i\in\mathcal{N}_k}
\sum_{j=1}^{J_i}
\Big[
&y_{ij}\log h_j(\mathbf{z}_{i}) \\
&+(1-y_{ij})\log\left(1-h_j(\mathbf{z}_{i})\right)
\Big].
\end{aligned}
\label{eq:discrete-bce-loss}
\end{equation}
where the model parameters are collected as $\boldsymbol{\Theta}=(\boldsymbol{\psi},\boldsymbol{\omega},\boldsymbol{\alpha})$, with $\boldsymbol{\psi}$ the temporal-encoder parameters, $\boldsymbol{\omega}$ the risk-function parameters, and $\boldsymbol{\alpha}=(\alpha_1,\ldots,\alpha_M)$ the interval-specific baseline parameters. This objective is a binary cross-entropy loss over the stacked system--interval observations. Since  each client evaluates Eq.~\eqref{eq:discrete-bce-loss} using only its local observations, the objective is client-separable, eliminating the need to construct global Cox risk sets.

The time grid determines the temporal resolution of the hazard model. Fine grids resolve short-term variation in risk but increase the number of interval-specific parameters
$\alpha_j$ and can produce intervals containing few or no failures. If an interval contains at-risk observations but no failures, the unregularized likelihood drives the corresponding baseline parameter toward $-\infty$. A coarser grid improves estimation stability by pooling events but may smooth short-term changes in risk. In the paper, two construction strategies are considered: equidistant discretization and a censoring-aware adaptive grid based on the Kaplan--Meier (KM) estimator \cite{rich2010practical}.

Under equidistant discretization, the grid is determined by a single width $\Delta_t$,
with boundaries $t_j=j\Delta_t$, $j=0,\ldots,M$.  With 
$I_j=(t_{j-1},t_j]$, the interval index of an observed time $T_i>0$ follows in closed
form,
\begin{equation}
J_i=\left\lceil \frac{T_i}{\Delta_t}\right\rceil.
\label{eq:quantize}
\end{equation}
Uniform spacing is suitable when observations, inspections, or maintenance decisions occur at regular time increments.

The second strategy derives the grid from the KM estimator. Let $u_1<u_2<\cdots$ denote the distinct failure times observed across all clients and define the pooled counts
\begin{equation}
d_\ell=\sum_{i\in\mathcal{N}}\mathbf{1}(T_i=u_\ell,\ \delta_i=1),
\qquad
r_\ell=\sum_{i\in\mathcal{N}}\mathbf{1}(T_i\geq u_\ell),
\label{eq:km-counts}
\end{equation}
where $\mathbf{1}(\cdot)$ is the indicator function,  $d_\ell$ and $r_\ell$ denote the
number of failures and the number of systems at risk at $u_\ell$, respectively. The survival function can be estimated nonparametrically using the KM estimator as
\begin{equation}
\widehat{S}_{\mathrm{KM}}(t)
=\prod_{u_\ell\leq t}\left(1-\frac{d_\ell}{r_\ell}\right).
\label{eq:km-estimator}
\end{equation}
Let $t_M$ denote the upper boundary of the modeled follow-up range. For
$j=1,\ldots,M-1$, the adaptive boundaries are defined as
\begin{equation}
t_j=\inf\left\{t:\widehat{S}_{\mathrm{KM}}(t)\leq
1-\frac{j}{M}\left[1-\widehat{S}_{\mathrm{KM}}(t_M)\right]\right\}.
\label{eq:km-cutpoint}
\end{equation}
This construction partitions the follow-up period according to the estimated failure-time distribution while accounting for right censoring. It allocates failure information more evenly across intervals and can improve the stability of the estimated interval-specific baseline parameters when failures are unevenly distributed over time. 

The KM-based grid, unlike the equidistant grid, depends on the failure-time distribution and must therefore be estimated from data. To avoid sharing individual failure times, the
central server first specifies a common fine reference grid
$
0=v_0<v_1<\cdots<v_B=t_M .
$
For each reference bin $b=1,\ldots,B$, client $k$ computes the local number of observed failures
$
d_b^{(k)}=\sum_{i\in\mathcal{N}_k}
\mathbf{1}\!\left(v_{b-1}<T_i\leq v_b,\ \delta_i=1\right)
$
and the local number of systems at risk at the start of the bin
$
r_b^{(k)}=\sum_{i\in\mathcal{N}_k}\mathbf{1}\!\left(T_i>v_{b-1}\right).
$
The server aggregates these counts as
\begin{equation}
d_b=\sum_{k=1}^{K}d_b^{(k)},
\qquad
r_b=\sum_{k=1}^{K}r_b^{(k)},
\end{equation}
and forms the binned KM estimate recursively from
$\widehat{S}_{\mathrm{KM}}^{\mathrm{bin}}(v_0)=1$ by
\begin{equation}
\widehat{S}_{\mathrm{KM}}^{\mathrm{bin}}(v_b)
=\widehat{S}_{\mathrm{KM}}^{\mathrm{bin}}(v_{b-1})
\left(1-\frac{d_b}{r_b}\right),
\label{eq:km-binned}
\end{equation}
with the convention that the factor equals one when $r_b=0$. The adaptive boundaries are
then obtained from Eq.~\eqref{eq:km-cutpoint} with $\widehat{S}_{\mathrm{KM}}$ replaced by
$\widehat{S}_{\mathrm{KM}}^{\mathrm{bin}}$ and broadcast to all clients in one preliminary communication round. The resulting grid remains fixed throughout training. This procedure discloses aggregated event and risk-set counts over the reference bins rather than individual failure times, and the bin width $v_b-v_{b-1}$ controls the granularity of that disclosure: a finer reference grid improves the approximation in Eq.~\eqref{eq:km-binned} but localizes failure times more precisely. 

\subsubsection{Attention-Based Temporal Representation Learning}
To incorporate longitudinal degradation information into the proposed model, the sensor sequence $\mathbf{X}_i$ is mapped to a covariate representation used in the interval-specific hazard. Static covariates and handcrafted summaries may not fully capture the nonlinear behavior, operating-regime changes, and long-term dependencies in multivariate sensor data \cite{zhou2012degradation}. We therefore use an attention-based temporal encoder to learn the representation directly from the sensor sequence.
For system $i$, let
$
\mathbf{X}_i
=
[\mathbf{x}_{i1},\mathbf{x}_{i2},\ldots,\mathbf{x}_{iL}]^{\top}
\in\mathbb{R}^{L\times p}
$
denote the most recent sensor window available at the prediction time, where $\mathbf{x}_{i\tau}\in\mathbb{R}^{p}$ is the sensor vector at step $\tau$ within the window, for $\tau=1,\ldots,L$, and $L$ is the window length. The window contains only measurements observed at or before the prediction time.
A Long Short-Term Memory (LSTM)  network first encodes the sequence and produces a hidden state at each time step,
\begin{equation}
\mathbf{h}_{i\tau}
=
\operatorname{LSTM}_{\boldsymbol{\psi}}
\left(
\mathbf{x}_{i\tau},\mathbf{h}_{i,\tau-1}
\right),
\qquad \tau=1,\ldots,L,
\label{eq:lstm-hidden}
\end{equation}
where $\mathbf{h}_{i\tau}\in\mathbb{R}^{d_h}$. Subsequently, additive attention \cite{bahdanau2014neural} assigns each hidden state the score
\begin{equation}
e_{i\tau}
=
\mathbf{v}^{\top}
\tanh\left(
\mathbf{W}_{a}\mathbf{h}_{i\tau}+\mathbf{b}_{a}
\right),
\label{eq:attention-score}
\end{equation}
with the corresponding normalized weight
\begin{equation}
a_{i\tau}
=
\frac{\exp(e_{i\tau})}
{\sum_{q=1}^{L}\exp(e_{iq})},
\qquad \tau=1,\ldots,L.
\label{eq:attention-weight}
\end{equation}
The resulting temporal representation is
\begin{equation}
\mathbf{z}_i
=
\sum_{\tau=1}^{L}
a_{i\tau}\mathbf{h}_{i\tau}.
\label{eq:attention-representation}
\end{equation}
Here, $\mathbf{W}{a}\in\mathbb{R}^{d_a\times d_h}$, $\mathbf{b}{a}\in\mathbb{R}^{d_a}$, and $\mathbf{v}\in\mathbb{R}^{d_a}$ are learned jointly with the LSTM parameters and they are  denoted by $\boldsymbol{\psi}$.
To further allow a nonlinear association between the learned representation and failure risk, we replace the linear predictor $\boldsymbol{\beta}^{\top}\mathbf{z}_i$ in Eq.~\eqref{eq:discrete-linear-hazard} with a scalar risk function $g_{\boldsymbol{\omega}}(\mathbf{z}_i)$, so that the interval-specific hazard becomes
\begin{equation}
h_j(\mathbf{z}_i)
=
1-
\exp\left\{
-\exp\left[
\alpha_j+g_{\boldsymbol{\omega}}(\mathbf{z}_i)
\right]
\right\}.
\label{eq:discrete-general-hazard}
\end{equation}
The function $g_{\boldsymbol{\omega}}$ contains no separate intercept, since any constant offset can be absorbed into the interval baseline parameters $\alpha_j$.
As new measurements arrive, a new window and representation are formed, so the predictions can be made  over time. 

\subsubsection{Federated Training and Prognostic Inference}
\begin{algorithm}[!t]
\caption{Federated Training and Prediction for the Proposed Discrete-Time Survival Model}
\label{alg:fed_survival}
\begin{algorithmic}[1]
\Require Client datasets $\{\mathcal{D}_k\}_{k=1}^{K}$, common discretization grid $\{t_j\}_{j=0}^{M}$ with interval widths $\Delta_j=t_j-t_{j-1}$, window length $L$
\Require Communication rounds $R$, local epochs $E$, learning rate $\eta$
\Ensure Trained global parameters $\boldsymbol{\Theta}^{(R)}$; local survival and RUL estimates
\State Server initializes
$\boldsymbol{\Theta}^{(0)}=
(\boldsymbol{\psi}^{(0)},\boldsymbol{\omega}^{(0)},\boldsymbol{\alpha}^{(0)})$.
\State Server broadcasts $\boldsymbol{\Theta}^{(0)}$ and $\{t_j\}_{j=0}^{M}$ to all clients.
\ForAll{clients $k=1,\ldots,K$ in parallel}
    \State Construct sliding-window sensor sequences of length $L$ from
    $\{\mathbf{X}_i\}_{i\in\mathcal{N}_k}$.
    \State Construct interval labels $y_{ij}$ from
    $\{(T_i,\delta_i)\}_{i\in\mathcal{N}_k}$ as in Eq.~\eqref{eq:stacked-label}.
\EndFor
\For{$r=0,1,\ldots,R-1$}
    \ForAll{clients $k=1,\ldots,K$ in parallel}
        \State Client $k$ receives $\boldsymbol{\Theta}^{(r)}$ and sets
        $\boldsymbol{\Theta}_k^{(r,0)}\gets\boldsymbol{\Theta}^{(r)}$.
        \For{$e=0,1,\ldots,E-1$}
            \State Encode each local sensor sequence to obtain
            $\mathbf{z}_i$.
            \State Compute interval-specific cloglog hazards
            $h_j(\mathbf{z}_i)$ via Eq.~\eqref{eq:discrete-general-hazard}.
            \State Evaluate the local discrete-time loss
            $\mathcal{L}_k(\boldsymbol{\Theta}_k^{(r,e)})$.
            \State Update the local parameters over one epoch of
            minibatch gradient steps:
            \[
            \boldsymbol{\Theta}_k^{(r,e+1)}
            \gets
            \operatorname{OptimizerStep}
            \!\left(
            \boldsymbol{\Theta}_k^{(r,e)},
            \nabla\mathcal{L}_k(\boldsymbol{\Theta}_k^{(r,e)}),
            \eta
            \right).
            \]
        \EndFor
        \State Set $\boldsymbol{\Theta}_k^{(r+1)}\gets\boldsymbol{\Theta}_k^{(r,E)}$ and send
        $\boldsymbol{\Theta}_k^{(r+1)}$ to the server.
    \EndFor
    \State Server aggregates the local parameters using FedAvg (Eq.~\eqref{eq:fedavg}):
    \[
    \boldsymbol{\Theta}^{(r+1)}
    \gets
    \sum_{k=1}^{K}\frac{N_k}{N}\,\boldsymbol{\Theta}_k^{(r+1)}.
    \]
\EndFor
\Statex \textit{Prediction (per client, using $\boldsymbol{\Theta}^{(R)}$):}
\State Compute interval hazards
$h_{ij}=h_j(\mathbf{z}_i)$.
\State Compute survival
$S_{ij}=\prod_{\ell=1}^{j}(1-h_{i\ell})$, with $S_{i0}=1$.
\State Estimate
$\widehat{\mathrm{RUL}}_{ij}
=
\dfrac{1}{S_{ij}}\sum_{\ell=j+1}^{M}
\Delta_{\ell}\,S_{i\ell}$, where $j=0$ denotes the prediction origin.
\end{algorithmic}
\end{algorithm}

The model is trained collaboratively across $K$ decentralized clients without pooling their sensor trajectories or time-to-event records. Recall that $\boldsymbol{\Theta}=(\boldsymbol{\psi},\boldsymbol{\omega},\boldsymbol{\alpha})$ collects the temporal-encoder, risk-function, and interval-baseline parameters. The global objective is
\begin{equation}
\widehat{\boldsymbol{\Theta}}
=
\arg\min_{\boldsymbol{\Theta}}
\mathcal{L}(\boldsymbol{\Theta})
=
\arg\min_{\boldsymbol{\Theta}}
\sum_{k=1}^{K}
\frac{N_k}{N}\,
\mathcal{L}_k(\boldsymbol{\Theta}),
\label{eq:federated-objective}
\end{equation}
where $\mathcal{L}_k$ is the local negative log-likelihood at client $k$ defined in Eq.~\eqref{eq:discrete-bce-loss}.

At communication round $r$, the server broadcasts the global parameters $\boldsymbol{\Theta}^{(r)}$ to the participating clients. Each client initializes its local model with these parameters and performs $E$ epochs of gradient-based optimization on $\mathcal{D}_k$,
\begin{equation}
\boldsymbol{\Theta}_k^{(r+1)}
=
\operatorname{LocalUpdate}
\left(
\boldsymbol{\Theta}^{(r)},\mathcal{D}_k,E,\eta
\right),
\label{eq:local-update}
\end{equation}
where $\eta$ is the learning rate. The server then forms the next global model through sample-size-weighted averaging,
\begin{equation}
\boldsymbol{\Theta}^{(r+1)}
=
\sum_{k=1}^{K}
\frac{N_k}{N}\,
\boldsymbol{\Theta}_k^{(r+1)}.
\label{eq:fedavg}
\end{equation}
These steps repeat for $R$ communication rounds, yielding the trained parameters $\boldsymbol{\Theta}^{(R)}$. The discretization grid is defined before training and shared across clients to ensure consistent interval-specific baseline parameters. Each client constructs its sensor windows and interval-level labels locally, while only model parameters are exchanged during training.

After training, the global trained model $\boldsymbol{\Theta}^{(R)}$ is used locally to estimate failure risk and RUL. Let $h_{ij}=h_j(\mathbf{z}_i)$ denote the predicted hazard for system $i$ in interval $I_j$, based on the temporal representation obtained from the sensor window available at the prediction time. The survival probability through the end of interval $I_j$ can be expressed as
\begin{equation}
S_{ij}
=
\Pr(T_i>t_j\mid\mathbf{z}_i)
=
\prod_{\ell=1}^{j}
\left(1-h_{i\ell}\right),
\label{eq:discrete-survival}
\end{equation}
and the cumulative probability of failure by $t_j$ is its complement,
\begin{equation}
F_{ij}
=
1-S_{ij}.
\label{eq:cumulative-failure}
\end{equation}
Conditional on survival through $t_j$, the RUL is
\begin{equation}
\mathrm{RUL}_{ij}
=
\mathbb{E}\left[
T_i-t_j
\mid
T_i>t_j,\mathbf{z}_i
\right],
\label{eq:rul-definition}
\end{equation}
which is approximated from the estimated survival curve over the remaining horizon as
\begin{equation}
\widehat{\mathrm{RUL}}_{ij}
=
\sum_{\ell=j+1}^{M}
\Delta_{\ell}\,
\frac{S_{i\ell}}{S_{ij}},
\qquad
\Delta_{\ell}=t_{\ell}-t_{\ell-1}.
\label{eq:rul-estimation}
\end{equation}
Here $S_{i\ell}/S_{ij}$ is the conditional probability of surviving beyond $t_{\ell}$ given survival beyond $t_j$. For an equidistant grid $\Delta_{\ell}=\Delta_t$; for the KM-based grid $\Delta_{\ell}$ accounts for unequal interval lengths. At the prediction origin, $j=0$ and $S_{i0}=1$.

As new sensor measurements become available, each client updates the input window and recomputes the hazard, survival, and RUL trajectories. The complete federated training and local prediction procedure is summarized in Algorithm~\ref{alg:fed_survival}.

\section{Case Study}
\label{sec:casestudy}

\subsection{Data Description}

\begin{figure}[t]
    \centering
    \includegraphics[width=0.95\columnwidth]{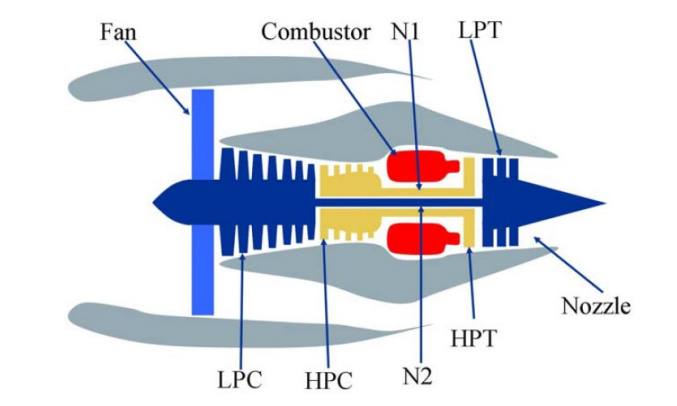}
    \caption{Turbofan-engine structure represented in the C-MAPSS simulation environment.}
    \label{fig_engine}
\end{figure}

The proposed framework is evaluated on the NASA Commercial Modular Aero-Propulsion System Simulation (C-MAPSS) dataset \cite{ramasso2014performance}, a widely used benchmark for prognostics and health management. C-MAPSS contains four run-to-failure turbofan-engine subsets, FD001--FD004, with different combinations of operating regimes and fault modes. These subsets represent progressively increasing levels of prognostic heterogeneity. Each engine starts in a healthy condition and is observed over consecutive flight cycles until failure in the training data or until a pre-failure truncation point in the test data. At each cycle, three operating settings and 21 sensor measurements are recorded from major engine components, including the fan, low-pressure compressor, high-pressure compressor, high-pressure turbine, and low-pressure turbine, as illustrated in Fig.~\ref{fig_engine}. The nominal subset characteristics and the numbers of test engines retained after window construction are summarized in Table~\ref{tab:dataset-characteristics}.

\subsection{Data Preprocessing}

\begin{table}[t]
\centering
\caption{Characteristics of the C-MAPSS Subsets}
\label{tab:dataset-characteristics}
\begin{tabular}{lcccc}
\toprule
Dataset & Regimes & Fault modes & Train engines & Test engines\textsuperscript{*} \\
\midrule
FD001 & 1 & 1 & 100 & 96 \\
FD002 & 6 & 1 & 260 & 244 \\
FD003 & 1 & 2 & 100 & 99 \\
FD004 & 6 & 2 & 249 & 232 \\
\bottomrule
\end{tabular}
\\[2pt]
\raggedright\footnotesize
\textsuperscript{*}Numbers retained after excluding test trajectories with 40 or fewer cycles, which cannot form a complete input window for $L=40$.
\end{table}

The raw trajectories are transformed into cycle-level RUL targets. For training engine $i$, let $T_i^{\mathrm{fail}}$ denote its final recorded cycle, corresponding to functional failure. Following common C-MAPSS practice, a piecewise-linear target is used to limit the influence of the early-life region, where degradation information is weak \cite{li2018remaining}. With the maximum RUL set to $R_{\max}=125$, the training target at cycle $t$ is
\[
    \mathrm{RUL}_i^{\mathrm{train}}(t)
    = \min\!\left(T_i^{\mathrm{fail}}-t,\;R_{\max}\right).
\]
Test trajectories are truncated before failure, and C-MAPSS provides the ground-truth RUL only at the final observed cycle. Let $T_{\mathrm{obs},i}$ be the last observed cycle of test engine $i$, and let $\mathrm{RUL}_i^{\mathrm{end}}$ be the corresponding provided RUL. The implied failure cycle is
\[
    T_i^{\mathrm{fail}}
    = T_{\mathrm{obs},i}+\mathrm{RUL}_i^{\mathrm{end}}.
\]
The RUL at any observed test cycle $t$ is then reconstructed as
\[
    \mathrm{RUL}_i^{\mathrm{test}}(t)
    = \min\!\left[
        \mathrm{RUL}_i^{\mathrm{end}}
        + \left(T_{\mathrm{obs},i}-t\right),\;
        R_{\max}
      \right].
\]
This reconstruction provides a target for every observed test cycle. Accordingly, the reported RUL metrics are computed over all valid test windows rather than only at the final observed cycle, enabling trajectory-level evaluation throughout the degradation process.

To account for operating-condition differences, the sensor and operating-setting channels are normalized using regime-specific training statistics \cite{wang2008similarity}. The three operating-setting variables are clustered by $k$-means into the known number of operating regimes: $K=1$ for FD001 and FD003 and $K=6$ for FD002 and FD004. Within each regime, every channel is standardized to zero mean and unit variance using the corresponding training-set mean and standard deviation. The same transformations are then applied to the test data. This regime-aware normalization reduces condition-induced offsets and scale differences in the raw measurements.

The normalized trajectories are segmented using a sliding window of length $L$, with each window forming a multivariate input sequence. Based on the sensitivity analysis in Section~\ref{window-size}, $L=40$ is used in the main experiments. Test trajectories containing fewer than 40 cycles cannot form a complete input window and are therefore excluded from evaluation. Accordingly, Table~\ref{tab:dataset-characteristics} reports the actual numbers of test engines included in the experiments.

\subsection{Evaluation Metrics}

We evaluate the proposed framework in terms of survival discrimination, calibration of predicted survival probabilities, and cycle-level RUL prediction accuracy.

\textit{1) Concordance index (C-index):}
The C-index assesses survival discrimination by measuring the concordance between predicted risk rankings and observed failure times. A higher value indicates that engines failing earlier are more consistently assigned higher predicted risk. It is calculated by
\begin{equation}
    C = P\!\left(r_i>r_j\mid T_i<T_j\right),
\end{equation}
where $r_i$ is the aggregate risk score and $T_i$ is the true remaining life of engine $i$. A value of 1 indicates perfect ranking, whereas 0.5 corresponds to random ranking.

\textit{2) Integrated Brier score (IBS):}
The IBS measures the squared discrepancy between the predicted survival probability $\hat{S}(t\mid\mathbf{z}_i)$ and the observed survival status $y_i(t)$ over the evaluation horizon:
\begin{equation}
    \mathrm{IBS}
    = \frac{1}{T_{\max}}
      \int_{0}^{T_{\max}}
      \frac{1}{N}\sum_{i=1}^{N}
      \left[\hat{S}(t\mid\mathbf{z}_i)-y_i(t)\right]^2dt.
\end{equation}
 Lower IBS values indicate better probabilistic accuracy.

\textit{3) Mean absolute error (MAE):}
The MAE measures the average absolute difference between the predicted and ground-truth RUL values across all valid test windows:
\begin{equation}
    \mathrm{MAE}
    = \frac{
        \sum_{i=1}^{N}\sum_{t=L}^{T_{\mathrm{obs},i}}
        \left|\widehat{\mathrm{RUL}}_i(t)-\mathrm{RUL}_i^{\mathrm{test}}(t)\right|
      }{
        \sum_{i=1}^{N}\left(T_{\mathrm{obs},i}-L+1\right)
      },
\end{equation}
where $ \widehat{\mathrm{RUL}}_i(t)$ denotes the RUL predicted for engine $i$ at cycle $t$.

\textit{4) Root mean squared error (RMSE):}
RMSE measures the square root of the average squared difference between the predicted and ground-truth RUL values across all valid test windows:
\begin{equation}
    \mathrm{RMSE}
    = \sqrt{
      \frac{
        \sum_{i=1}^{N}\sum_{t=L}^{T_{\mathrm{obs},i}}
        \left(\widehat{\mathrm{RUL}}_i(t)-\mathrm{RUL}_i^{\mathrm{test}}(t)\right)^2
      }{
        \sum_{i=1}^{N}\left(T_{\mathrm{obs},i}-L+1\right)
      }}.
\end{equation}

\subsection{Results and Discussion}

We evaluate the proposed framework in two stages. First, FD001 is used as a controlled setting to select the temporal window length and event-time discretization scheme under federated training. The selected settings are then fixed and applied uniformly across FD001--FD004 and across all training paradigms, ensuring that subsequent differences reflect dataset heterogeneity and training strategy rather than paradigm-specific hyperparameter tuning. Second, using these fixed settings, we assess sensitivity to the link function and compare federated, centralized, and local-only training across all four subsets.

\subsubsection{Selection of Window Length and Discretization Scheme} 
\label{window-size}
\begin{figure}[t]
    \centering
    \includegraphics[width=0.95\columnwidth]{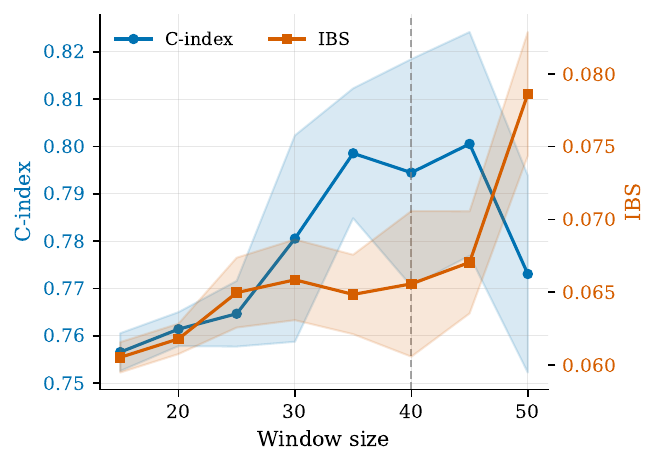}
    \caption{Sensitivity of the proposed model to temporal window length on FD001. }
    \label{fig:window_size}
\vspace{-6mm}
\end{figure}

We first examine sensitivity to the temporal window length by varying
$L \in \{15,20,25,30,35,40,45,50\}$.
Each configuration is evaluated over five runs with different random seeds, and the results are reported as the mean $\pm$ standard deviation. The corresponding results are presented in 
Table~\ref{tab:window-sweep} and Fig.~\ref{fig:window_size}. It is observed that the mean
C-index generally increases with $L$, reaching approximately 0.80 for
$L=35$--45. RMSE also decreases over this range relative to the
shorter windows. IBS varies moderately for $L\leq45$ but increases from
0.0671 at $L=45$ to 0.0786 at $L=50$, indicating deterioration in
calibration for the longest window.
Within the stable region $L=35$--45, the differences in the mean
metrics are small relative to the observed seed-to-seed variability. We
therefore select $L=40$ as a representative interior point that
balances discrimination, calibration, and RUL prediction accuracy while
avoiding the performance degradation observed at $L=50$.

\begin{table}[t]
\centering
\caption{Window-Length Sensitivity on FD001 (Mean $\pm$ Standard Deviation Over Five Seeds)}
\label{tab:window-sweep}
\begin{tabular}{cccc}
\toprule
Window size & C-index $\uparrow$ & IBS $\downarrow$ & RMSE $\downarrow$ \\
\midrule
15 & $0.757 \pm 0.004$ & $0.0605 \pm 0.0010$ & $19.19 \pm 0.45$ \\
20 & $0.761 \pm 0.004$ & $0.0618 \pm 0.0010$ & $18.62 \pm 0.58$ \\
25 & $0.765 \pm 0.007$ & $0.0650 \pm 0.0024$ & $19.21 \pm 1.00$ \\
30 & $0.781 \pm 0.022$ & $0.0659 \pm 0.0028$ & $18.04 \pm 1.28$ \\
35 & $0.799 \pm 0.014$ & $0.0648 \pm 0.0027$ & $17.25 \pm 0.76$ \\
40 & $0.794 \pm 0.024$ & $0.0656 \pm 0.0050$ & $17.71 \pm 1.65$ \\
45 & $0.801 \pm 0.024$ & $0.0671 \pm 0.0029$ & $17.42 \pm 0.89$ \\
50 & $0.773 \pm 0.021$ & $0.0786 \pm 0.0043$ & $19.62 \pm 0.87$ \\
\bottomrule
\end{tabular}
\\[2pt]
\end{table}

\subsubsection{Event-Time Discretization}

Using $L=40$, we next compare equidistant and KM-based event-time
discretization. A target of 50 intervals is specified for both
strategies to ensure a comparable initial temporal resolution. For
equidistant discretization, this yields 50 uniformly spaced intervals.
For KM-based discretization, repeated empirical quantiles are merged,
resulting in 37 distinct effective intervals. As shown in Table~\ref{tab:discretization-scheme}, KM-based
discretization reduces the mean IBS from 0.0694 to 0.0656 and the mean
RMSE from 18.46 to 17.70, corresponding to relative reductions of
approximately 5.5\% and 4.1\%, respectively. The mean C-index also
increases slightly from 0.785 to 0.787. Thus, the KM-based grid
achieves better calibration and lower RUL prediction error using a
more compact event-time representation, while maintaining comparable
discrimination. Based on these results, KM-based discretization is
used in the remaining experiments.

\begin{table}[t]
\centering
\caption{Comparison of Event-Time Discretization Schemes on FD001 ($L=40$; Mean $\pm$ Standard Deviation Over Five Seeds)}
\label{tab:discretization-scheme}
\setlength{\tabcolsep}{4pt}
\begin{tabular}{lcccc}
\toprule
Scheme & Bins & C-index $\uparrow$ & IBS $\downarrow$ & RMSE $\downarrow$ \\
\midrule
KM (default) & 37 & $0.787 \pm 0.024$ & $0.0656 \pm 0.0050$ & $17.70 \pm 1.64$ \\
Equidistant & 50 & $0.785 \pm 0.011$ & $0.0694 \pm 0.0034$ & $18.46 \pm 0.63$ \\
\bottomrule
\end{tabular}
\end{table}

\subsubsection{Effect of the Link Function}
\begin{table*}[!t]
\centering
\caption{Comparison of Cloglog and Logit Link Functions Using Paired Wilcoxon Signed-Rank Tests ($L=40$)}
\label{tab:link-validation}
\begin{tabular}{llccccc}
\toprule
Dataset & Level & Metric & $n$ & Logit & Cloglog & $p$-value \\
\midrule
\multirow{4}{*}{FD001}
 & Federated & C-index & 5  & 0.7959 & 0.7871 & 0.313 \\
 & Federated & IBS     & 5  & 0.0643 & 0.0656 & 0.313 \\
 & Local     & C-index & 10 & 0.6859 & 0.6855 & 0.846 \\
 & Local     & IBS     & 10 & 0.1270 & 0.1243 & 0.375 \\
\midrule
\multirow{4}{*}{FD002}
 & Federated & C-index & 5  & 0.8463 & 0.8483 & 0.625 \\
 & Federated & IBS     & 5  & 0.0649 & 0.0638 & 0.313 \\
 & Local     & C-index & 10 & 0.7412 & 0.7428 & 0.846 \\
 & Local     & IBS     & 10 & 0.1269 & 0.1224 & 0.020* \\
\midrule
\multirow{4}{*}{FD003}
 & Federated & C-index & 5  & 0.7372 & 0.7432 & 0.625 \\
 & Federated & IBS     & 5  & 0.0558 & 0.0572 & 0.438 \\
 & Local     & C-index & 10 & 0.6747 & 0.6739 & 1.000 \\
 & Local     & IBS     & 10 & 0.1253 & 0.1246 & 0.846 \\
\midrule
\multirow{4}{*}{FD004}
 & Federated & C-index & 5  & 0.7941 & 0.7577 & 0.063 \\
 & Federated & IBS     & 5  & 0.0567 & 0.0635 & 0.125 \\
 & Local     & C-index & 10 & 0.7286 & 0.7308 & 0.625 \\
 & Local     & IBS     & 10 & 0.0982 & 0.0955 & 0.010* \\
\bottomrule
\end{tabular}
\end{table*}

Using the selected window length and event-time discretization scheme, we next evaluate the sensitivity of the framework to the link function. The cloglog formulation is compared with a logit-link benchmark using (L=40) and KM-based event-time discretization. As discussed in Section~\ref{sec:Discretization}, the cloglog link arises from discretizing the continuous-time Cox PH model, whereas the logit link corresponds to a discrete-time proportional-odds formulation. This comparison therefore serves as a robustness analysis, assessing whether the predictive performance of the framework depends strongly on the link-function assumption. Table~\ref{tab:link-validation} reports the corresponding C-index and IBS values, together with paired Wilcoxon signed-rank tests at the federated and local levels.

The two link functions yield broadly comparable predictive performance. No federated-level comparison and no C-index comparison reaches the nominal $p<0.05$ level. At the local level, cloglog yields lower mean IBS on FD002 and FD004, with unadjusted $p$-values of 0.020 and 0.010, respectively. However, neither difference remains significant after Holm correction across the 16 tests. These findings indicate that the main predictive results are not driven by the choice between the cloglog and logit links and do not establish a consistent advantage for either formulation. The cloglog link is retained in the remaining experiments as it preserves the proportional-hazards interpretation of the continuous-time Cox model.

\subsubsection{Comparison of Training Strategies}

\begin{figure}[t]
    \centering
    \includegraphics[width=0.95\columnwidth]{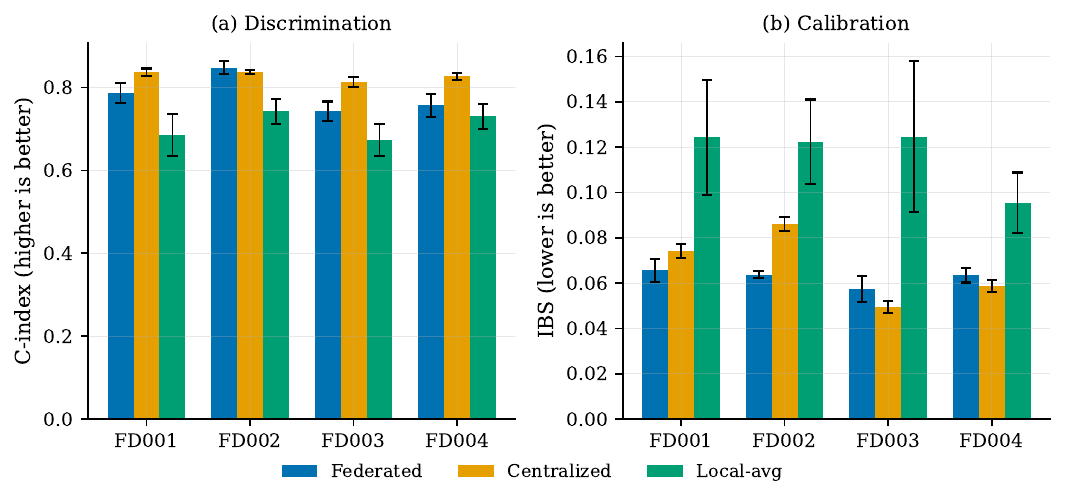}
    \caption{Comparison of federated, centralized, and local-average training across the four C-MAPSS subsets. Federation consistently improves on local-only training, while its relationship with centralized training varies with subset heterogeneity.}
    \label{fig:main_comparison}
\end{figure}

\begin{figure}[t]
    \centering
    \includegraphics[width=0.95\columnwidth]{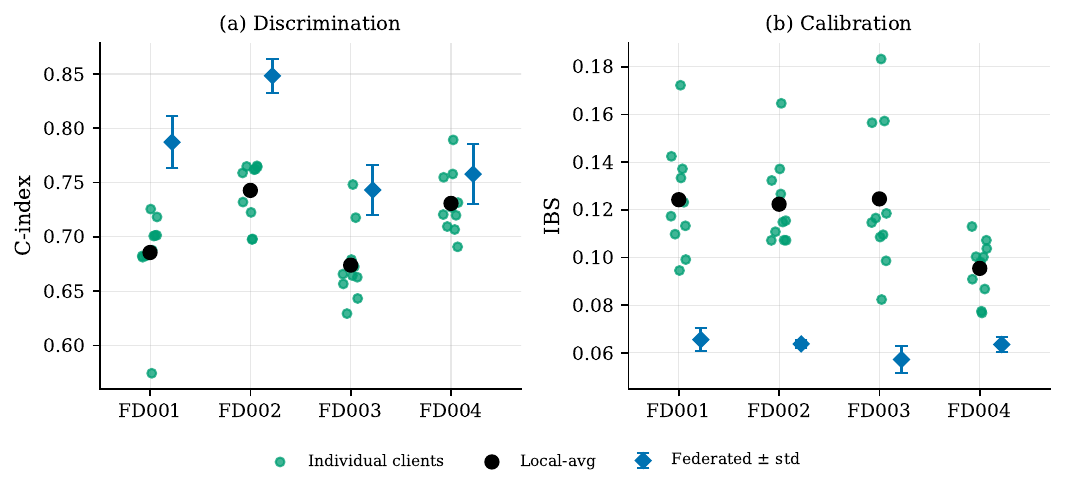}
    \caption{Distribution of local-client C-index values across the ten clients, with the federated result shown as a reference. The spread illustrates the variability and instability of isolated local training.}
    \label{fig:per_client_variance}
\vspace{-6mm}
\end{figure}

Using the design choices selected on FD001, we evaluate the proposed framework across all four C-MAPSS subsets, FD001--FD004, which represent different operating conditions and fault modes.  For each subset, the training engines are partitioned by engine identity across (K=10) disjoint simulated clients, with all windows from the same engine assigned to a single client. This preserves complete engine trajectories and represents a multisite setting in which engine data cannot be pooled. Federated training aggregates client updates, while the local-only baseline trains one independent model per client without aggregation. Centralized training uses the same engines pooled into a single training set. All three paradigms are evaluated on the same held-out test engines, ensuring that the reported differences reflect the training strategy rather than differences in training or test data.
Table~\ref{tab:main-comparison} and Fig.~\ref{fig:main_comparison}  compare the performance of federated, centralized, and local-only training, while Fig.~\ref{fig:per_client_variance} presents the distribution of C-index values across local clients.

\begin{table*}[t]
\centering
\caption{Comparison of Federated, Centralized, and Local-Only Training (Mean $\pm$ Standard Deviation Over Five Seeds; Local Summaries Computed Across Ten Clients)}
\label{tab:main-comparison}
\begin{tabular}{llcccc}
\toprule
Dataset & Method & C-index $\uparrow$ & IBS $\downarrow$ & MAE $\downarrow$ & RMSE $\downarrow$ \\
\midrule
\multirow{5}{*}{FD001}
 & Federated (proposed) & $0.787 \pm 0.024$ & $0.0656 \pm 0.0050$ & $11.42 \pm 0.71$ & $17.70 \pm 1.64$ \\
 & Centralized & $0.838 \pm 0.009$ & $0.0742 \pm 0.0031$ & $11.11 \pm 0.44$ & $17.25 \pm 0.35$ \\
 & Local-avg & $0.686 \pm 0.050$ & $0.1243 \pm 0.0253$ & $18.51 \pm 3.87$ & $26.88 \pm 4.31$ \\
 & Local-best & $0.726 \pm 0.042$ & $0.0946 \pm 0.0113$ & $15.11 \pm 1.89$ & $23.00 \pm 2.54$ \\
 & Local-worst & $0.574 \pm 0.068$ & $0.1723 \pm 0.0281$ & $27.15 \pm 5.13$ & $36.54 \pm 5.88$ \\
\midrule
\multirow{5}{*}{FD002}
 & Federated (proposed) & $0.848 \pm 0.006$ & $0.0638 \pm 0.0017$ & $11.05 \pm 0.38$ & $18.02 \pm 0.49$ \\
 & Centralized & $0.837 \pm 0.005$ & $0.0861 \pm 0.0030$ & $12.52 \pm 0.28$ & $19.01 \pm 0.44$ \\
 & Local-avg & $0.743 \pm 0.030$ & $0.1224 \pm 0.0186$ & $17.34 \pm 2.32$ & $24.94 \pm 2.53$ \\
 & Local-best & $0.765 \pm 0.019$ & $0.1155 \pm 0.0105$ & $16.35 \pm 1.12$ & $23.73 \pm 1.43$ \\
 & Local-worst & $0.698 \pm 0.010$ & $0.1267 \pm 0.0057$ & $17.55 \pm 0.58$ & $26.00 \pm 0.41$ \\
\midrule
\multirow{5}{*}{FD003}
 & Federated (proposed) & $0.743 \pm 0.023$ & $0.0572 \pm 0.0058$ & $9.33 \pm 0.94$ & $17.64 \pm 1.65$ \\
 & Centralized & $0.814 \pm 0.011$ & $0.0493 \pm 0.0027$ & $7.18 \pm 0.32$ & $13.43 \pm 0.46$ \\
 & Local-avg & $0.674 \pm 0.039$ & $0.1246 \pm 0.0333$ & $17.11 \pm 4.32$ & $27.89 \pm 5.39$ \\
 & Local-best & $0.748 \pm 0.014$ & $0.0824 \pm 0.0025$ & $11.27 \pm 0.33$ & $21.75 \pm 0.63$ \\
 & Local-worst & $0.629 \pm 0.010$ & $0.1166 \pm 0.0113$ & $16.45 \pm 1.23$ & $26.94 \pm 1.43$ \\
\midrule
\multirow{5}{*}{FD004}
 & Federated (proposed) & $0.758 \pm 0.028$ & $0.0635 \pm 0.0033$ & $10.49 \pm 0.55$ & $20.44 \pm 1.04$ \\
 & Centralized & $0.827 \pm 0.007$ & $0.0588 \pm 0.0027$ & $8.24 \pm 0.21$ & $16.00 \pm 0.29$ \\
 & Local-avg & $0.731 \pm 0.030$ & $0.0955 \pm 0.0133$ & $12.92 \pm 1.69$ & $22.47 \pm 2.08$ \\
 & Local-best & $0.789 \pm 0.011$ & $0.0767 \pm 0.0019$ & $10.63 \pm 0.16$ & $20.04 \pm 0.61$ \\
 & Local-worst & $0.691 \pm 0.008$ & $0.1072 \pm 0.0051$ & $14.35 \pm 0.77$ & $24.33 \pm 0.99$ \\
\bottomrule
\end{tabular}
\end{table*}

\begin{table}[!htbp]
\centering
\caption{Performance preservation relative to centralized training (\%).}
\label{tab:preservation}
\resizebox{\columnwidth}{!}{%
\begin{tabular}{llcccc}
\toprule
Dataset & Method & $P_{C}$ & $P_{\mathrm{IBS}}$ & $P_{\mathrm{MAE}}$ & $P_{\mathrm{RMSE}}$ \\
\midrule
\multirow{2}{*}{FD001}
 & Federated & 93.9 & 113.1 & 97.3 & 97.5 \\
 & Local-avg & 81.9 & 59.7 & 60.0 & 64.2 \\
\midrule
\multirow{2}{*}{FD002}
 & Federated & 101.3 & 134.9 & 113.3 & 105.5 \\
 & Local-avg & 88.8 & 70.3 & 72.2 & 76.2 \\
\midrule
\multirow{2}{*}{FD003}
 & Federated & 91.3 & 86.2 & 77.0 & 76.2 \\
 & Local-avg & 82.8 & 39.6 & 42.0 & 48.1 \\
\midrule
\multirow{2}{*}{FD004}
 & Federated & 91.7 & 92.6 & 78.6 & 78.2 \\
 & Local-avg & 88.4 & 61.6 & 63.8 & 71.2 \\
\bottomrule
\end{tabular}%
}
\\[2pt]
\raggedright\footnotesize
For C-index, $P_C=100\times C_{\mathrm{Method}}/C_{\mathrm{Centralized}}$.
For IBS, MAE, and RMSE, $P_E=100\times E_{\mathrm{Centralized}}/E_{\mathrm{Method}}$.
Thus, larger values consistently indicate better performance, and $100\%$ denotes parity with centralized training.
\end{table}

From Table~\ref{tab:main-comparison}  and Fig.~\ref{fig:main_comparison}  we can observe that the proposed federated model outperforms the Local-avg baseline across all four subsets and all four metrics. Relative to Local-avg, federated training increases the C-index by 0.027--0.105, decreases IBS by 0.0320--0.0674, decreases MAE by 2.43--7.78 cycles, and decreases RMSE by 2.03--10.25 cycles. The largest improvements are observed on FD001--FD003. Although the gains are smaller on FD004, which contains six operating conditions and two fault modes, federated training continues to improve all four metrics relative to Local-avg.
The client-level dispersion in Fig.~\ref{fig:per_client_variance} further illustrates the instability of isolated local training. On FD001, for example, the Local-worst result has a C-index of 0.574 and an RMSE of 36.54 cycles, compared with 0.787 and 17.70 cycles, respectively, under federated training. These results show that federated aggregation not only improves average predictive performance but also reduces the risk of poor outcomes caused by limited client-specific data.

\begin{figure*}[t]
    \centering
    \includegraphics[width=0.85\textwidth]{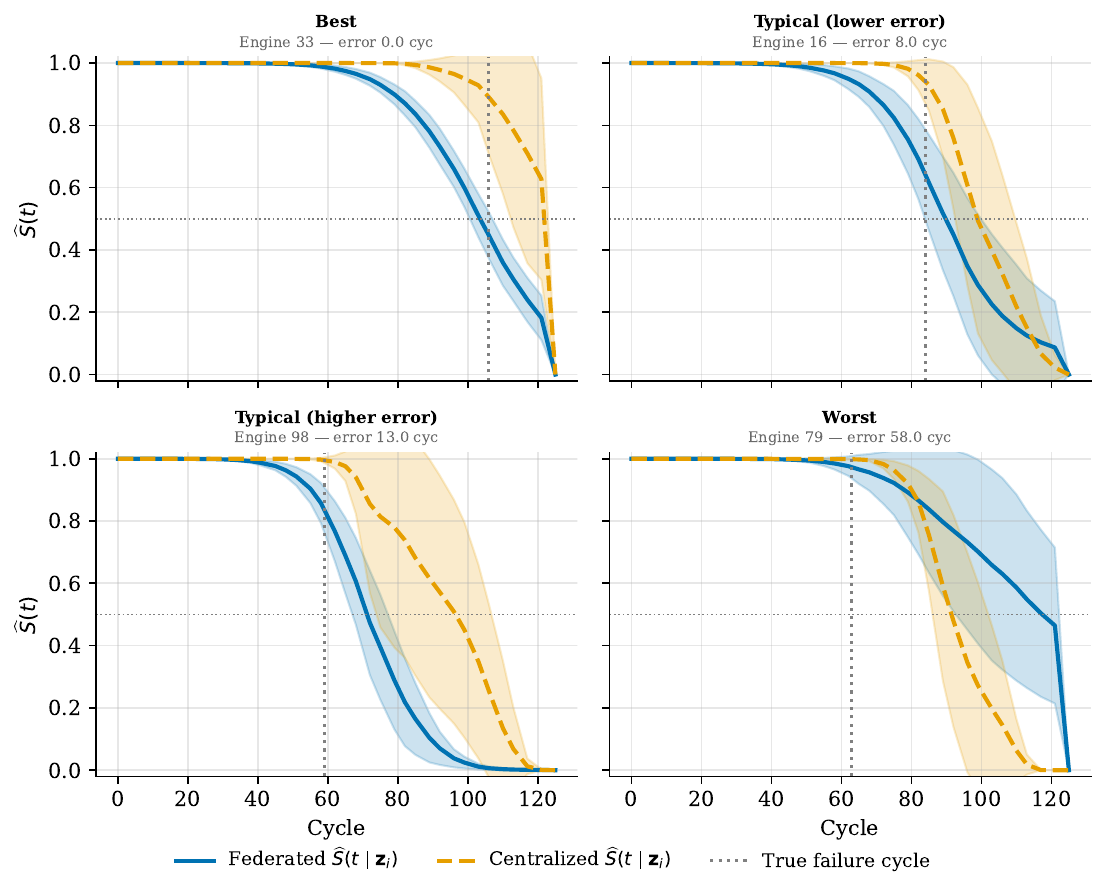}
\caption{Predicted survival curves for four FD001 test engines selected to span the absolute RUL-error distribution of the federated model, from the best- to the worst-performing case. Solid and dashed curves denote federated and centralized predictions, respectively. Shaded bands indicate $\pm 1$ standard deviation across five training seeds, and dotted vertical lines mark the corresponding ground-truth failure cycles.}
    \label{fig:example_survival_curves}
\end{figure*}

Compared with centralized training, federated performance is dataset- and metric-dependent. On FD001, the federated model shows lower discrimination but better calibration, while MAE and RMSE remain similar. On FD002, federated training performs better across all four metrics. In contrast, centralized training retains a clear advantage on FD003 and FD004, particularly in discrimination and RUL prediction accuracy. The greater fault-mode heterogeneity of these subsets may reduce the effectiveness of global parameter aggregation. Nevertheless, the federated model continues to outperform Local-avg across every metric, demonstrating that cross-client collaboration remains beneficial even when centralized performance is not attained.

Table~\ref{tab:preservation} summarizes the performance preservation of federated and local-only training relative to centralized training. Across all four C-MAPSS subsets, federated training preserves a substantially larger fraction of centralized performance than local-only training. On FD001 and FD002, the federated model achieves performance close to or better than centralized training for most metrics, indicating that collaborative learning can effectively recover the benefit of pooled data without direct data sharing. On the more challenging FD003 and FD004 subsets, which involve multiple failure modes, the gap between federated and centralized training becomes more pronounced, particularly for MAE and RMSE. Nevertheless, federated training remains consistently stronger than local-only training, demonstrating the value of cross-client collaboration for survival-based prognostics under data-local constraints.

Aggregate metrics can obscure substantial engine-level heterogeneity.
Fig.~\ref{fig:example_survival_curves} therefore compares federated and
centralized predictions for four selected FD001 engines: the best-performing case (Engine~33), the worst-performing case (Engine~79), and two representative intermediate cases (Engines~16 and
98). For Engines~33, 16, and 98, the federated survival curve declines near the true failure cycle, whereas the centralized curve remains
close to $\widehat{S}=1$ for longer and crosses $\widehat{S}=0.5$ later, indicating delayed failure-risk assessment. The opposite pattern
is observed for Engine~79, where the federated model predicts failure substantially too late, while the centralized curve more closely
reflects the true failure timing. The seed-to-seed uncertainty bands are generally widest during the transition from high to low survival,
indicating greater variability when degradation becomes evident. Beyond point RUL error, these results demonstrate an important
advantage of survival modeling: the predicted curves characterize the
full evolution and uncertainty of failure risk over time rather than providing only a single RUL estimate.

Overall, the case study demonstrates that the proposed federated survival framework consistently improves upon local-only training without pooling client data and achieves strong prognostic performance across fleets with diverse operating conditions and fault modes.

\section{Conclusion and Future Work}
\label{sec:conclusion}

This study developed a federated longitudinal--survival framework for collaborative system prognostics using decentralized sensor data. By integrating temporal representation learning with discrete-time Cox proportional-hazards modeling, the framework estimates interval-specific failure risks, survival probabilities, and RUL without pooling raw sensor trajectories or individual time-to-event records. The client-separable likelihood enables federated optimization while retaining the proportional-hazards interpretation through the cloglog link. Experiments on the four C-MAPSS subsets show that collaborative training consistently improves performance over isolated local models and achieves dataset-dependent performance relative to centralized training.

Future work will investigate client-specific and hierarchical model components to better accommodate heterogeneous operating conditions and failure mechanisms. Uncertainty quantification and calibration methods will also be incorporated to provide prediction intervals and reliability guarantees for RUL estimates. Additional directions include adaptive event-time discretization, partial client participation, communication-efficient optimization, secure aggregation, and differential privacy. Finally, evaluation on additional industrial datasets with censoring, irregular sampling, and realistic cross-site heterogeneity will be conducted to assess the generalizability and deployment readiness of the proposed framework.


\bibliography{bib2}

@article{lei2018machinery,
  title={Machinery health prognostics: A systematic review from data acquisition to RUL prediction},
  author={Lei, Yaguo and Li, Naipeng and Guo, Liang and Li, Ningbo and Yan, Tao and Lin, Jing},
  journal={Mechanical systems and signal processing},
  volume={104},
  pages={799--834},
  year={2018},
  publisher={Elsevier}
}

@article{bahdanau2014neural,
  title={Neural machine translation by jointly learning to align and translate},
  author={Bahdanau, Dzmitry and Cho, Kyunghyun and Bengio, Yoshua},
  journal={arXiv preprint arXiv:1409.0473},
  year={2014}
}

@article{andersen1982cox,
  title={Cox's regression model for counting processes: a large sample study},
  author={Andersen, Per Kragh and Gill, Richard D},
  journal={The annals of statistics},
  pages={1100--1120},
  year={1982},
  publisher={JSTOR}
}

@article{cox1975partial,
  title={Partial likelihood},
  author={Cox, David R},
  journal={Biometrika},
  volume={62},
  number={2},
  pages={269--276},
  year={1975},
  publisher={Oxford University Press}
}

@inproceedings{wang2008similarity,
  title={A similarity-based prognostics approach for remaining useful life estimation of engineered systems},
  author={Wang, Tianyi and Yu, Jianbo and Siegel, David and Lee, Jay},
  booktitle={2008 international conference on prognostics and health management},
  pages={1--6},
  year={2008},
  organization={IEEE}
}

@article{li2018remaining,
  title={Remaining useful life estimation in prognostics using deep convolution neural networks},
  author={Li, Xiang and Ding, Qian and Sun, Jian-Qiao},
  journal={Reliability Engineering \& System Safety},
  volume={172},
  pages={1--11},
  year={2018},
  publisher={Elsevier}
}

@book{kalbfleisch2002statistical,
  title={The statistical analysis of failure time data},
  author={Kalbfleisch, John D and Prentice, Ross L},
  year={2002},
  publisher={John Wiley \& Sons}
}

@article{allison1982discrete,
  title={Discrete-time methods for the analysis of event histories},
  author={Allison, Paul D},
  journal={Sociological methodology},
  volume={13},
  pages={61--98},
  year={1982},
  publisher={JSTOR}
}

@article{prentice1978regression,
  title={Regression analysis of grouped survival data with application to breast cancer data},
  author={Prentice, Ross L and Gloeckler, Lynn A},
  journal={Biometrics},
  pages={57--67},
  year={1978},
  publisher={JSTOR}
}

@incollection{zhang2022federated,
  title={A federated cox model with non-proportional hazards},
  author={Zhang, D Kai and Toni, Francesca and Williams, Matthew},
  booktitle={Multimodal AI in healthcare: A paradigm shift in health intelligence},
  pages={171--185},
  year={2022},
  publisher={Springer}
}

@article{andreux2020federated,
  title={Federated Survival Analysis with Discrete-Time Cox Models},
  author={Andreux, Mathieu and Manoel, Andre and Romagnoni, Alberto and Wack, Charles and others},
  journal={arXiv preprint arXiv:2006.08997},
  year={2020}
}

@inproceedings{sateesh2016deep,
  title={Deep convolutional neural network based regression approach for estimation of remaining useful life},
  author={Sateesh Babu, Giduthuri and Zhao, Peilin and Li, Xiao-Li},
  booktitle={International conference on database systems for advanced applications},
  pages={214--228},
  year={2016},
  organization={Springer}
}

@article{ramasso2014performance,
  title={Performance benchmarking and analysis of prognostic methods for cmapss datasets.},
  author={Ramasso, Emmanuel and Saxena, Abhinav},
  journal={International Journal of Prognostics and Health Management},
  volume={5},
  number={2},
  pages={1--15},
  year={2014}
}

@article{kairouz2021advances,
  title={Advances and open problems in federated learning},
  author={Kairouz, Peter and McMahan, H Brendan and Avent, Brendan and Bellet, Aur{\'e}lien and Bennis, Mehdi and Bhagoji, Arjun Nitin and Bonawitz, Kallista and Charles, Zachary and Cormode, Graham and Cummings, Rachel and others},
  journal={Foundations and trends{\textregistered} in machine learning},
  volume={14},
  number={1--2},
  pages={1--210},
  year={2021},
  publisher={Now Publishers, Inc.}
}

@article{zhang2023privacy,
  title={Privacy-preserving and sensor-fused framework for prognostic \& health management in leased manufacturing system},
  author={Zhang, Kaigan and Xia, Tangbin and Wang, Dong and Chen, Genliang and Pan, Ershun and Xi, Lifeng},
  journal={Mechanical Systems and Signal Processing},
  volume={184},
  pages={109666},
  year={2023},
  publisher={Elsevier}
}

@article{yue2021joint,
  title={Joint models for event prediction from time series and survival data},
  author={Yue, Xubo and Kontar, Raed Al},
  journal={Technometrics},
  volume={63},
  number={4},
  pages={477--486},
  year={2021},
  publisher={Taylor \& Francis}
}

@article{yan2018functional,
  title={Functional principal components analysis on moving time windows of longitudinal data: dynamic prediction of times to event},
  author={Yan, Fangrong and Lin, Xiao and Li, Ruosha and Huang, Xuelin},
  journal={Journal of the Royal Statistical Society Series C: Applied Statistics},
  volume={67},
  number={4},
  pages={961--978},
  year={2018},
  publisher={Oxford University Press}
}

@article{zhou2014remaining,
  title={Remaining useful life prediction of individual units subject to hard failure},
  author={Zhou, Qiang and Son, Junbo and Zhou, Shiyu and Mao, Xiaofeng and Salman, Mutasim},
  journal={IIE transactions},
  volume={46},
  number={10},
  pages={1017--1030},
  year={2014},
  publisher={Taylor \& Francis}
}

@article{hu2020condition,
  title={Condition-based maintenance planning for systems subject to dependent soft and hard failures},
  author={Hu, Jiawen and Sun, Qiuzhuang and Ye, Zhi-Sheng},
  journal={ieee Transactions on Reliability},
  volume={70},
  number={4},
  pages={1468--1480},
  year={2020},
  publisher={IEEE}
}

@article{wen2022recent,
  title={Recent advances and trends of predictive maintenance from data-driven machine prognostics perspective},
  author={Wen, Yuxin and Rahman, Md Fashiar and Xu, Honglun and Tseng, Tzu-Liang Bill},
  journal={Measurement},
  volume={187},
  pages={110276},
  year={2022},
  publisher={Elsevier}
}

@article{wen2018degradation,
  title={Degradation modeling and RUL prediction using Wiener process subject to multiple change points and unit heterogeneity},
  author={Wen, Yuxin and Wu, Jianguo and Das, Devashish and Tseng, Tzu-Liang Bill},
  journal={Reliability Engineering \& System Safety},
  volume={176},
  pages={113--124},
  year={2018},
  publisher={Elsevier}
}

@article{zhu2019survey,
  title={A survey of predictive maintenance: Systems, purposes and approaches},
  author={Zhu, Tianwen and Ran, Yongyi and Zhou, Xin and Wen, Yonggang},
  journal={arXiv preprint arXiv:1912.07383},
  year={2019}
}

@article{zio2022prognostics,
  title={Prognostics and Health Management (PHM): Where are we and where do we (need to) go in theory and practice},
  author={Zio, Enrico},
  journal={Reliability Engineering \& System Safety},
  volume={218},
  pages={108119},
  year={2022},
  publisher={Elsevier}
}

@article{wen2023neural,
  title={A neural-network-based proportional hazard model for IoT signal fusion and failure prediction},
  author={Wen, Yuxin and Guo, Xinxing and Son, Junbo and Wu, Jianguo},
  journal={IISE Transactions},
  volume={55},
  number={4},
  pages={377--391},
  year={2023},
  publisher={Taylor \& Francis}
}

@article{dhada_empirical_2020,
    series = {4th {IFAC} {Workshop} on {Advanced} {Maintenance} {Engineering}, {Services} and {Technologies} - {AMEST} 2020},
    title = {Empirical {Convergence} {Analysis} of {Federated} {Averaging} for {Failure} {Prognosis}⁎},
    volume = {53},
    issn = {2405-8963},
    url = {https://www.sciencedirect.com/science/article/pii/S2405896320302081},
    doi = {10.1016/j.ifacol.2020.11.058},
    journal = {IFAC-PapersOnLine},
    author = {Dhada, Maharshi and Jain, Amit Kumar and Parlikad, Ajith Kumar},
    month = jan,
    year = {2020},
    pages = {360--365},
}

@inproceedings{mcmahan2017communication,
  title={Communication-Efficient Learning of Deep Networks from Decentralized Data},
  author={McMahan, H. Brendan and Moore, Eider and Ramage, Daniel and Hampson, Seth and y Arcas, Blaise A.},
  booktitle={Proceedings of the 20th International Conference on Artificial Intelligence and Statistics},
  year={2017}
}

@article{katzman2018deepsurv,
  title={DeepSurv: personalized treatment recommender system using a Cox proportional hazards deep neural network},
  author={Katzman, Jared L. and Shaham, Uri and Bates, Jonathan and Cloninger, Alexander and Jiang, Tingting and Kluger, Yuval},
  journal={BMC Medical Research Methodology},
  volume={18},
  number={1},
  pages={24},
  year={2018},
  publisher={BioMed Central}
}

@inproceedings{lee2018deephit,
title={DeepHit: A deep learning approach to survival analysis with competing risks},
author={Lee, Changhee and Zame, William R. and Yoon, Jinsung and van der Schaar, Mihaela},
booktitle={Proceedings of the AAAI Conference on Artificial Intelligence},
volume={32},
number={1},
year={2018}
}

@article{zhang2022dast,
title={Dual Aspect Self-Attention Based on Transformer for Remaining Useful Life Prediction},
author={Zhang, Zhizheng and Song, Wen and Li, Qiqiang},
journal={IEEE Transactions on Instrumentation and Measurement},
volume={71},
pages={1--11},
year={2022},
publisher={IEEE}
}

@article{arunan2023federated,
title={A Federated Learning-Based Industrial Health Prognostics for Heterogeneous Edge Devices Using Matched Feature Extraction},
author={Arunan, Anushiya and Qin, Yan and Li, Xiaoli and Yuen, Chau},
journal={arXiv preprint arXiv:2305.07854},
year={2023}
}

@inproceedings{li2020fedprox,
  title={Federated Optimization in Heterogeneous Networks},
  author={Li, Tian and Sahu, Anit Kumar and Zaheer, Manzil and Sanjabi, Maziar and Talwalkar, Ameet and Smith, Virginia},
  booktitle={Proceedings of Machine Learning and Systems},
  volume={2},
  pages={429--450},
  year={2020}
}

@inproceedings{karimireddy2020scaffold,
title={SCAFFOLD: Stochastic Controlled Averaging for Federated Learning},
author={Karimireddy, Sai Praneeth and Kale, Satyen and Mohri, Mehryar and Reddi, Sashank J. and Stich, Sebastian U. and Suresh, Ananda Theertha},
booktitle={Proceedings of the 37th International Conference on Machine Learning},
pages={5132--5143},
year={2020}
}

@article{brumm2025joint,
  title={Joint modeling of degradation signals and time-to-event data for the prediction of remaining useful life},
  author={Brumm, Sebastian and Linstead, Erik and Chen, Junde and Balakrishnan, Narayanaswamy and Wen, Yuxin},
  journal={Quality and Reliability Engineering International},
  volume={41},
  number={2},
  pages={607--624},
  year={2025},
  publisher={Wiley Online Library}
}

@article{rich2010practical,
  title={A practical guide to understanding Kaplan-Meier curves},
  author={Rich, Jason T and Neely, J Gail and Paniello, Randal C and Voelker, Courtney CJ and Nussenbaum, Brian and Wang, Eric W},
  journal={Otolaryngology—Head and Neck Surgery},
  volume={143},
  number={3},
  pages={331--336},
  year={2010},
  publisher={SAGE Publications Sage CA: Los Angeles, CA}
}

@article{zhou2012degradation,
  title={Degradation modeling and monitoring of truncated degradation signals},
  author={Zhou, Rensheng and Gebraeel, Nagi and Serban, Nicoleta},
  journal={IIE Transactions},
  volume={44},
  number={9},
  pages={793--803},
  year={2012},
  publisher={Taylor \& Francis}
}

\end{document}